\documentclass[sigconf,screen, nonacm, authorversion]{acmart}
\copyrightyear{2023}
\acmYear{2023}
\setcopyright{acmlicensed}\acmConference[ICVIP 2023]{2023 the 7th International Conference on Video and Image Processing (ICVIP)}{December 14--17, 2023}{Kyoto, Japan}
\acmBooktitle{2023 the 7th International Conference on Video and Image Processing (ICVIP) (ICVIP 2023), December 14--17, 2023, Kyoto, Japan}
\acmDOI{10.1145/3639390.3639392}
\acmISBN{979-8-4007-0938-8/23/12}
\usepackage{graphicx}
\usepackage{pifont}
\usepackage{multirow}
\usepackage{subfig}
\newcommand{\cmark}{\ding{51}}
\newcommand{\xmark}{\ding{55}}
\usepackage{pgfplots}
\graphicspath{{./images/}}

\usepackage[capitalize]{cleveref}
\crefname{section}{Sec.}{Secs.}
\Crefname{section}{Section}{Sections}
\Crefname{table}{Table}{Tables}
\crefname{table}{Tab.}{Tabs.}
\def\BibTeX{{\rm B\kern-.05em{\sc i\kern-.025em b}\kern-.08em
		T\kern-.1667em\lower.7ex\hbox{E}\kern-.125emX}}
\newcommand{\natten}{$\mathcal{N}ATTEN$ }

\AtBeginDocument{%
	\providecommand\BibTeX{{%
			\normalfont B\kern-0.5em{\scshape i\kern-0.25em b}\kern-0.8em\TeX}}}

\copyrightyear{2023}

\begin{document}
	
	\title[NAC-TCN: Temporal Convolutional Networks with Causal Dilated Neighborhood Attention]{NAC-TCN: Temporal Convolutional Networks with Causal Dilated Neighborhood Attention for Emotion Understanding}
	
	\author{Alexander Mehta}
	\email{alexandermehta@outlook.com}
		\affiliation{%
	\institution{Independent Researcher}  \country{USA} }
	
	\author{William Yang}
	\email{yangwill@seas.upenn.edu}
	
	\affiliation{\institution{University of Pennsylvania}  \country{USA}}

	\renewcommand{\shortauthors}{Mehta and Yang}
	
	\begin{abstract}
	In the task of emotion recognition from videos, a key improvement has been to focus on emotions over time rather than a single frame. There are many architectures to address this task such as GRUs, LSTMs, Self-Attention, Transformers, and Temporal Convolutional Networks (TCNs). However, these methods suffer from high memory usage, large amounts of operations, or poor gradients. We propose a method known as \textbf{N}eighborhood\textbf{ A}ttention with \textbf{C}onvolutions \textbf{TCN} (NAC-TCN) which incorporates the benefits of attention and Temporal Convolutional Networks while ensuring that causal relationships are understood which results in a reduction in computation and memory cost. We accomplish this by introducing a causal version of Dilated Neighborhood Attention while incorporating it with convolutions.  Our model achieves comparable, better, or \textit{state-of-the-art} performance over TCNs, TCAN, LSTMs, and GRUs while requiring fewer parameters on standard emotion recognition datasets. We publish our code online for easy reproducibility and use in other projects -- \href{https://github.com/alexmehta/NAC-TCN-TCNs-with-Causal-NA}{ \color{blue}  Github Link}.  
		\end{abstract}



		\maketitle
		
\begin{figure}[!htbp]
	\centering
	\includegraphics[scale=0.25]{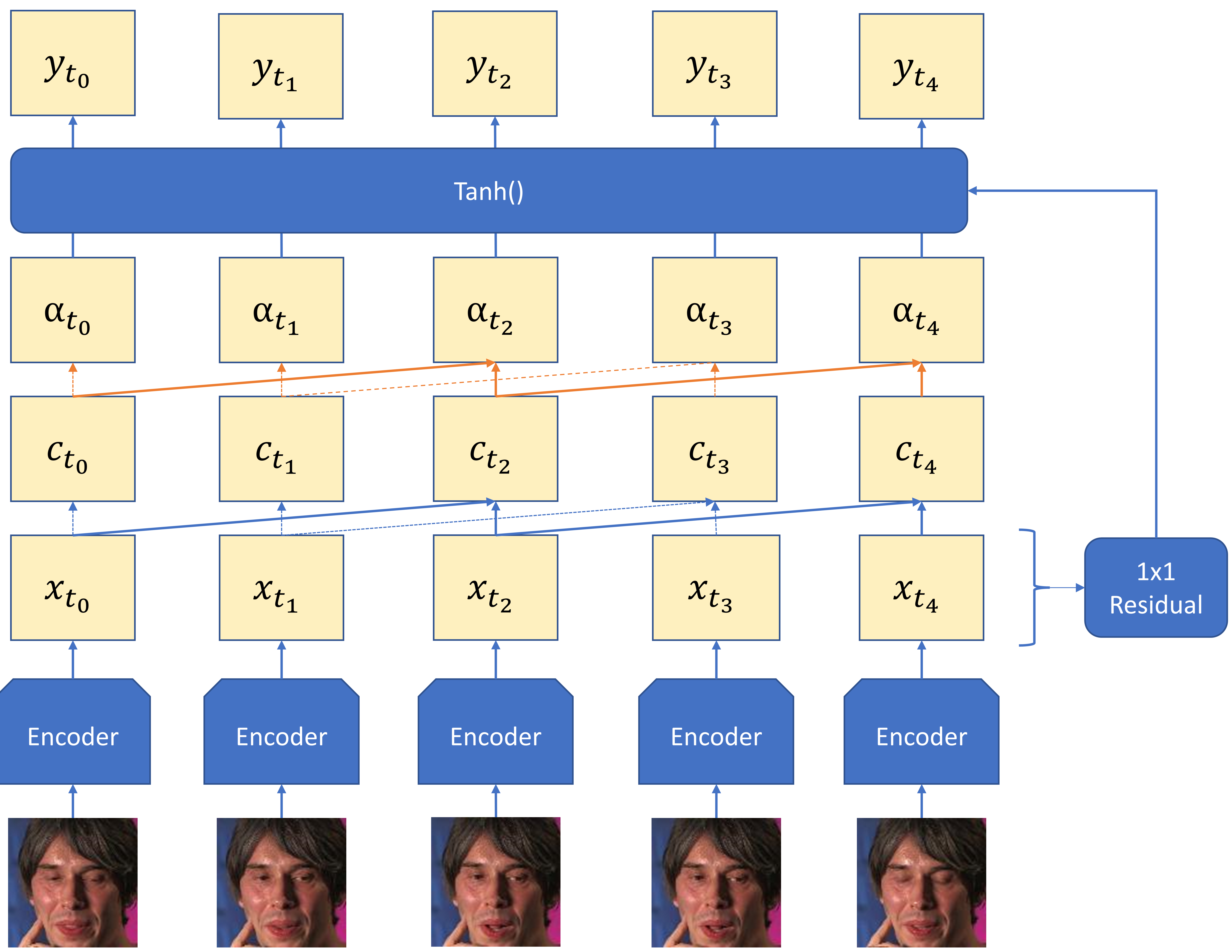}
	\caption{The NAC-TCN combines Dilated Temporal Convolutions with Dilated Neighborhood Attention to better capture temporal relationships in video inputs through contextual weighting using Dilated Neighborhood Attention. Our proposed architecture achieves \textbf{better performance with smaller model size}. }
	\label{fig:nac-tcn-temporal-block}
\end{figure}
\section{Introduction}
\label{sec:intro}

The study of emotion recognition has gained significant importance due to its widespread applications in various disciplines, including Human-Computer Interaction. Socially assistive robots, medical diagnosis of disorders such as PTSD \cite{schultebraucks2022deep}, and software usability testing \cite{kolakowska2014emotion} in particular rely heavily on accurate emotion recognition. For instance, a better understanding of human emotions enables socially assistive robots to comprehend and respond appropriately to various scenarios, leading to effective assistance \cite{Abdollahi_2022}. As a result, the development of advanced emotion recognition techniques holds immense potential in revolutionizing various aspects of our lives.

A common issue with the task of emotion recognition and understanding is lack of data. This is commonly due to the fact that annotation is a non-trivial task. This means that architectures must be expressive enough without large data sizes. 
The classification of emotions from video inputs is a task that has been extensively studied in the field of computer vision. A common approach to this problem involves performing classification directly on individual frames using convolutional networks\cite{saroop2021facial}. However, this approach ignores the temporal aspect of emotion, which is critical to its accurate recognition. Humans exhibit emotions over a period of time, and considering this temporal aspect can lead to significant improvements in performance through contextual understanding \cite{DBLP:journals/corr/abs-1905-02947}.

To incorporate temporal information into emotion recognition from video inputs, various techniques have been proposed, including Recurrent Neural Networks (RNNs) \cite{rangulov2020emotion} and Transformers \cite{TCNTransformer,GUPTA20201527,carion2020endtoend}. While GRUs and LSTMs are effective in modeling temporal dependencies, they suffer from slow training times, unstable training, and high computational costs. This is because their gradients flow through time rather than in parallel \cite{jozefowicz2015empirical}. Transformers, on the other hand, typically suffer have a large parameter size and require more operations due to their self-attention mechanism attending to a large receptive field. Commonly the two models are combined for optimal performance \cite{9190996}.

An alternative solution that has gained popularity is the Temporal Convolutional Network (TCN)  \cite{yu2016multiscale, BaiTCN2018, oord2016wavenet}. The TCN allows for the modeling of temporal dependencies in video tasks and time-series tasks similar to RNNs and LSTMs, but with more stable gradients and higher efficiency at large receptive fields due to parallelized computation of gradients \cite{DBLP:journals/corr/LeaFVRH16, BaiTCN2018}. The receptive field of the TCN can be easily adjusted with the number of layers, kernel size, and dilation factor. The TCN is a promising method for video tasks due to its efficiency benefits while understanding temporal dependencies \cite{9251902} for emotional understanding. 

Temporal Convolutional Networks do pose performance issues in regard to it's understanding complex relationships in the short and long term of a sequence and more irregular sequences \cite{moor2019early}. Models attempt to address this, but fall short in terms of model size and performance due to the large size of self attention or auxiliary models \cite{moor2019early,bai2018trellis, TCNTransformer}. TCAN attempts to intertwine attention layers  \cite{hao2020temporal} which leads to larger models. 

In this paper we introduce a new Temporal Convolutional Network known as NAC-TCN which addresses concerns about complex temporal relationships while maintaining or improving the benefits of TCNs -- stable gradients and computational efficiency.


\subsection{Related Works}
\subsubsection{Recurrent Networks} \label{sec:rec}
Long-Short Term Memory Units (LSTMs) are a response to the hard to train nature of RNNs \cite{rnn_train}. RNNs store a hidden state that can be used in order to represent all prior knowledge. Though this may be a strong representation, there are often training issues in long sequences. LSTMs include a forget gate that can allow for data to not be stored in the cells memory unit state. The LSTM has an input, output, and forget gate. LSTM models suffer from a large amount of operations per cell. Gated Recurrent Units \cite{cho2014learning} attempt to solve this by not holding a memory unit, instead only having update and forget gates. Despite the less complex structure, it has similar performance to LSTMs \cite{chung2014empirical}. 
\subsubsection{Attention and Transformer}
\label{sec:attn}
Self Attention \cite{cho2014learning,vaswani2017attention} was introduced for language tasks. Instead of simply holding a weight and bias, self attention focuses on parts of the input and weights sequential inputs based on a query, key, and value matrix. These attention models can be used in conjunction with recurrent models for better performance (\cref{sec:rec}). 

Transformers use layered attention with encoders and decoders \cite{vaswani2017attention, devlin2019bert}. This has been commonly used in NLP tasks, but recent advances have moved towards it's application in computer vision \cite{Arnab_2021_ICCV}.

These 2 methods represent a divergence in machine learning -- use of classical recurrent methods or a move to the more computationally heavy transformer. In this paper, we hope to show a third path that can incorporate the benefits of both while alleviating their pitfalls. 

\section{Background}

\subsection{Temporal Convolutional Networks}
\label{sec:TCN}
The Temporal Convolutional Network \cite{yu2016multiscale, oord2016wavenet,BaiTCN2018} is a convolutional representation of temporal data. It contains 2 commonly used parts, the main casual convolution network (\cref{ss:tcn}) and dilated convolution (\cref{ss:dc}) in order to create a Dilated Temporal Convolutional Network. 

\subsubsection{Dilated Convolution}
\label{ss:dc}
The dilation (à trous Convolution) \cite{yu2016multiscale} allows for a model to have a larger receptive field without increasing parameters. Dilated convolution is achieved by introducing "holes" between the points addressed by the kernel, resulting in a larger receptive field. The term ``gaps'' will be used to refer to any method of expanding a kernel through gaps to widen the receptive field. 
\subsubsection{Dilated Temporal Convolutional Network}
\label{ss:tcn}
Introduced in WaveNet \cite{oord2016wavenet}, a Dilated Temporal Convolutional Network is a temporal network model that computes timesteps in parallel rather than sequentially. This fundamentally alters how the model addresses backpropagation through time by performing backpropogation for all time steps at once rather than following a temporal gradient flow. A casual convolution is used in order to prevent leakage from the past into future steps. 

\begin{figure}[!htbp]
	
	\includegraphics[width=\linewidth, scale=0.027]{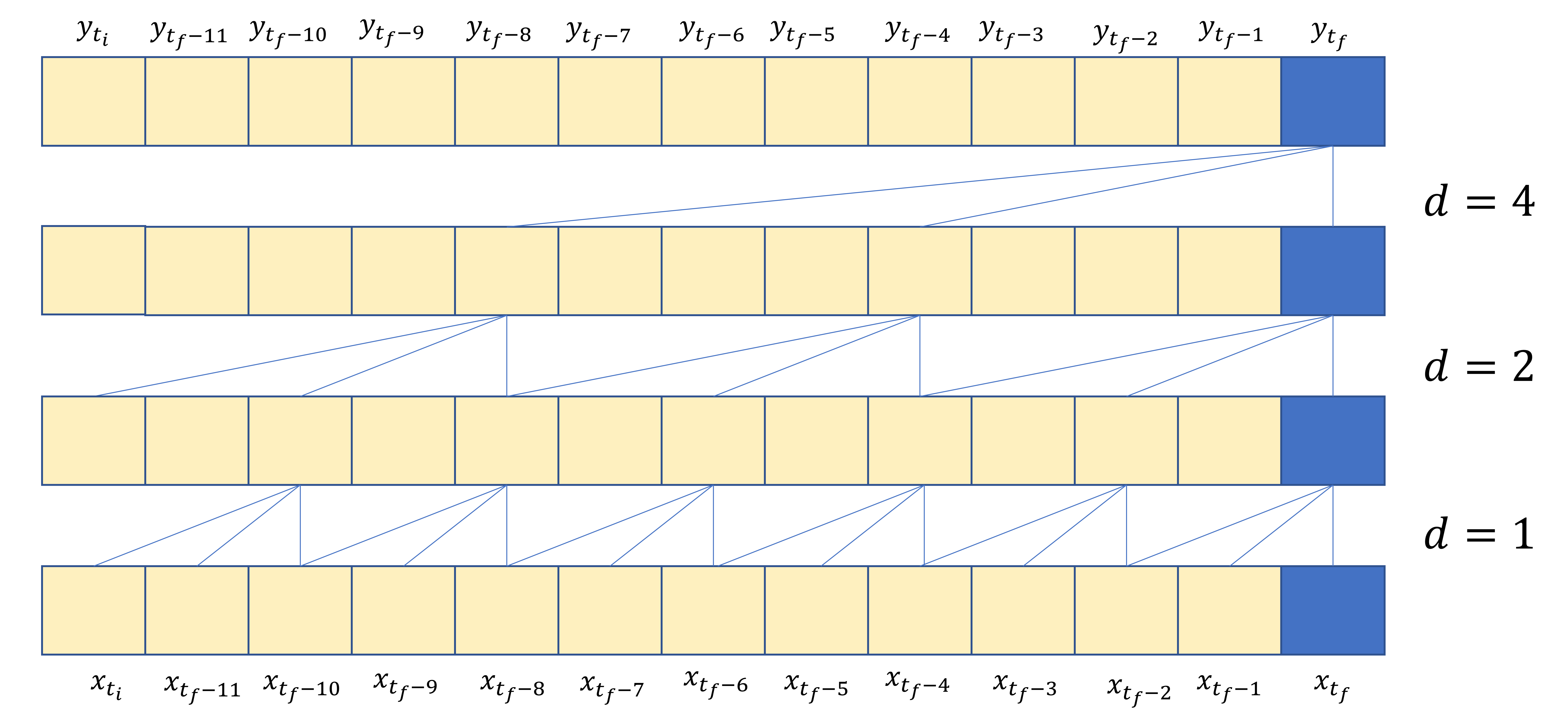}
	\caption{TCN architecture}
	\label{fig:tcn_block_figure}
\end{figure}

A TCN layer can be described as follows for an input sequence $x$, dilation $d$, length $i$, dilated convolution $\ast_d$, and filter $f$
\begin{align}
	F(x_t) = (x \ast_d f)(t) = \sum_{n=0}^{i-1} f(t) \cdot x_{s-d*i}.
	\label{Equation1}
\end{align}
A dilated convolution is used in order to allow a network to understand time steps from previous steps efficiently and exponentially increase the receptive field \cite{yu2016multiscale}. Without dilated convolutions, TCNs would have a linear receptive field to prior steps. With dilations, the receptive field to previous timesteps (frames) can be calculated as 
\begin{align}
	RF(n,d,k) = 1 + \sum_{i=0}^{n-1} d^i(k-1), 
\end{align}
where $d$ is the dilation factor, $k$ is the kernel size, and $n$ is the number of hidden layers. Commonly, a dilation factor of 2 is used in order to achieve an exponential receptive field \cite{BaiTCN2018}. 

Dilated Temporal Convolutional Networks (TCNs) allow for a large amount of temporal data to be processed with low computation through a large receptive field. TCNs allow for parallel computation, a large receptive field, and helps avoid vanishing or exploding gradients due to its backpropagation not being parallel to the temporal sequence, but rather perpendicular. Dilated TCNs have achieved impressive results replicating the long-term memory understanding of other architectures like LSTMs and RNNs such as the copying memory task \cite{BaiTCN2018}. The TCN has also been adopted for action segmentation achieving state-of-the-art results on action detection \cite{lea2017temporal}. TCNs have also been explored in emotion analysis, achieving results above those of LSTMs and RNNs on emotion based tasks \cite{TCNTransformer}. TCNs commonly consist of temporal blocks, which are made up of two convolutional layers that are stacked on top of each other. The purpose of stacking these layers is to ensure that the input data is first scaled to the expected size and then passed through a convolutional layer of the output size.


\subsection{Neighborhood Attention}
\label{sec:DiNA}
Neighborhood Attention (NA) is an attention method that utilizes a sliding window technique similar to a convolution which views the time series at incriments like a convolution instead of all at once such as self-attention. This is similar to methods such as the SWIN transformer \cite{liu2021swin,hassani2022neighborhood} but the main difference comes from how NA allows for overlapping sgements, a method showed to improve performance by ensuring translation equivariance over similar methods \cite{hassani2022neighborhood, hassani2022dilated}. NA was introduced in order to address poor efficiency of self-attention and sliding window techniques by using a tiled algorithm and efficient CUDA kernels published in the  \natten  library \cite{natten}. 

\paragraph{Dilated Neighborhood Attention (DiNA)}is a method introduced to further address the performance of attention \cite{hassani2022dilated}. This dilated transformer works similar to dilated (also known as à trous) convolutions \cite{yu2016multiscale}. This improves performance beyond Neighborhood attention by attending to a higher receptive field in less operations than a normal transformer. When $d$ is a dilation value and $k$ is a neighborhood (kernel) size, DiNA reduces time complexity of self-attention from $O(n^2d)$ to $O(ndk)$.

\subsubsection{TCAN} TCAN \cite{hao2020temporal} is a TCN-based model that intertwines attention to maintain receptive field while providing an attention mechanism. This model has seen improvements over the TCN on language datasets. Although the method has seen improvements over the TCN, it leads to a significant parameter increase and it doesn't preserve the casual nature of the TCN, meaning that information flows freely between layers, and loses the temporal property due to leakage.

\section{Methods}
\begin{figure*}[!ht]
	\centering
	\subfloat[\centering $\alpha$ represents the Attention. Convolutions and Attention for timesteps other than $t_f$ are omitted for simplicity. ]{{		\includegraphics[scale=0.35]{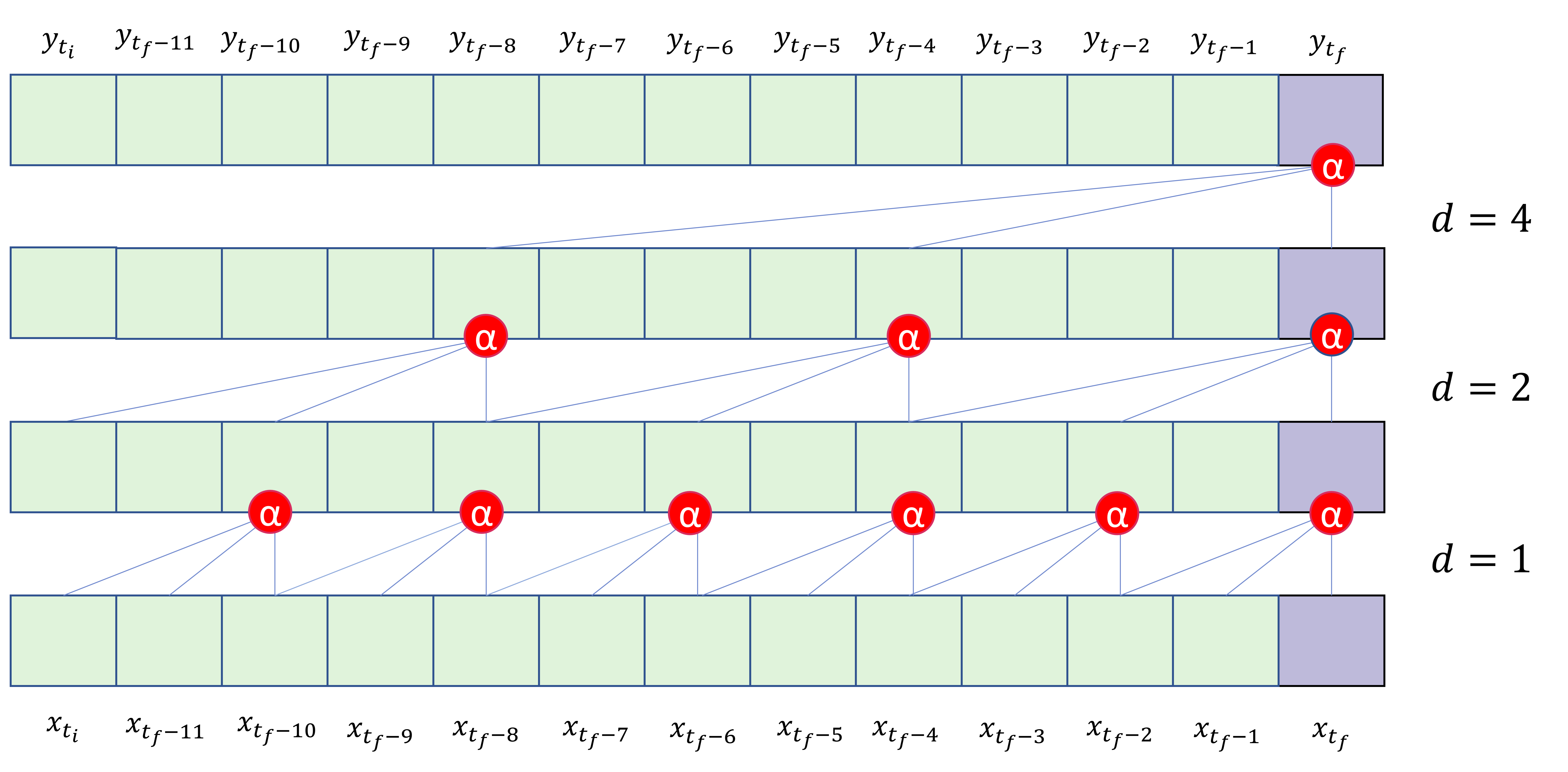}}}\label{NACATCN_diagram_1} 
	\qquad
	\subfloat[\centering NAC-TCN Temporal Block]{{		\includegraphics[scale=0.35]{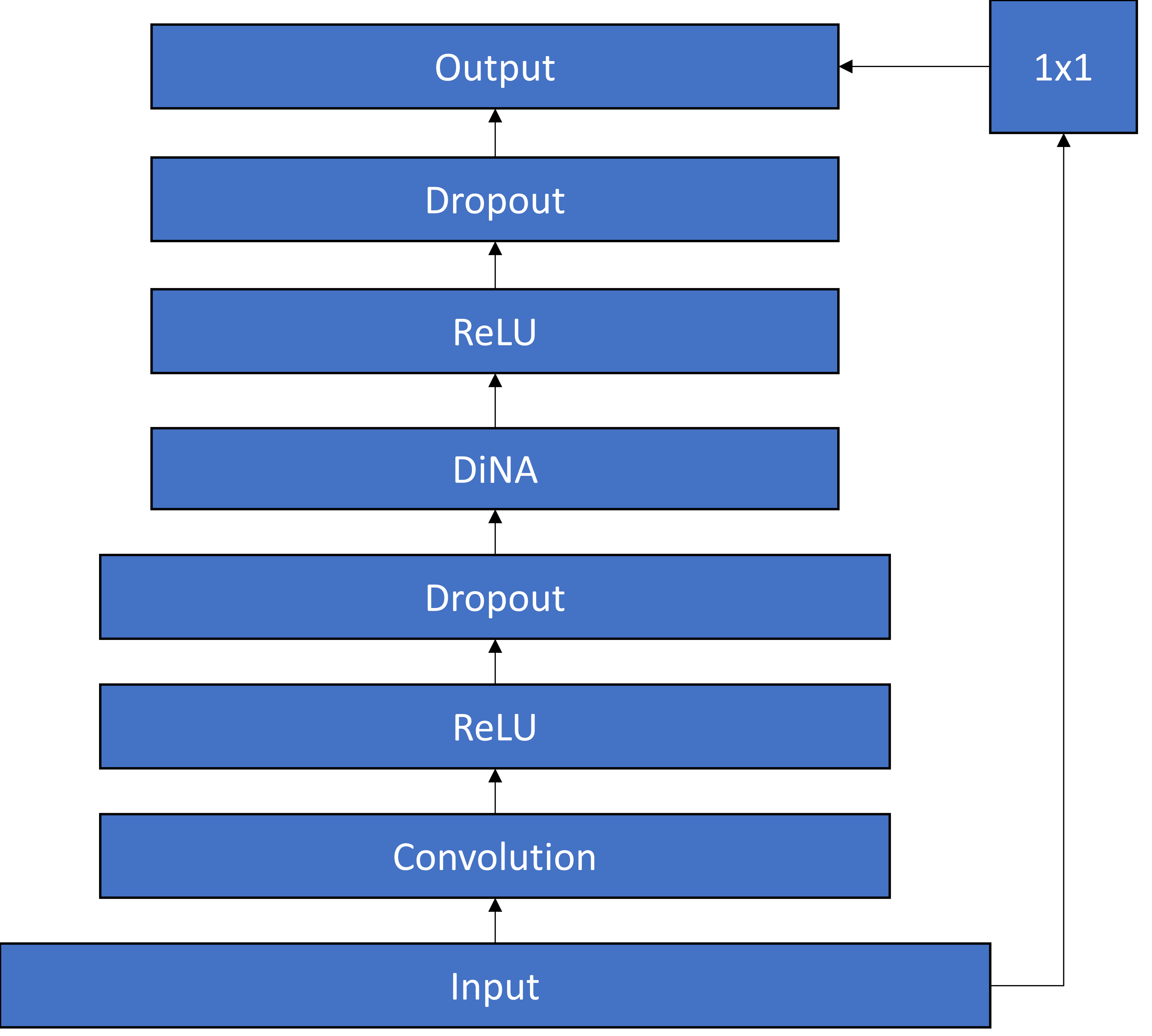}}} \label{NACATCN_diagram_2}
\end{figure*}

To enhance the representation of temporal dependencies and their importance in emotional understanding, we propose a method that extends TCNs by incorporating the attention features of Neighborhood Attention while maintaining causality. We introduce our proposed archetecture that achieves this along with memory and runtime benefits in this section. 
\subsection{NAC-TCN Formulation}
The \textbf{N}eighborhood \textbf{A}ttention with feature extracting \textbf{C}onvolutions TCN (NAC-TCN), is a deep learning based approach that utilizes Dilated Neighborhood Attention to enforce causality and combines convolutional operations and self-attention. Our proposed NAC-TCN method incorporates neighborhood self-attention layers within Temporal Blocks with 1D Convolutional Layers to allow the TCN to identify the most important frames through Neighborhood Attention and create local filters through 1D Convolutions. A combination of convolution and attention layers has been shown to produce improved results \cite{xu2021coscale}. Our method makes use of Dilated Neighborhood Attention \cite{hassani2022dilated,hassani2022neighborhood,walton2022stylenat}, and shifting inputs to maintain causality. 1x1 convolutions \cite{lin2014network} are added on the input of each temporal block in order to ensure that residual connections have the same tensor shape in a similar fashion to the original TCN. The use of Dilated Neighborhood Attention not only keeps causality in the TCN, but also reduces operations and parameters.

\subsection{Temporal Block}
A NAC-TCN temporal block is parametrized by its kernel size $k$, dilation value $d$, input $x$, time step $t$, convolution $f_k$ and the DiNA operation (\cref{eq:DiNAten}) and can be described as 
\begin{align}
	F(x_t) = (x \ast_d f_k \ast_d \text{DiNA}_{k}^{d})(t).
	\label{NAT-TCN equation}	
\end{align}
In between each convolution, an activation (ReLU) is applied and followed by a 1D Spatial Dropout Layer which allows for feature independence between channels of the model \cite{tompson2015efficient}. This reflects the primary diagram in \cref{NACATCN_diagram_2} which shows the Temporal Block structure where a Convolution is followed by Dropout and ReLU with a 1x1 convolution as described in \cref{sec:residual}.

\subsubsection{Motivation for Convolution and Neighborhood Attention Stacking}
We wish to create high performing low operational cost models. Adding convolutions achieves this twofold: being able to reduce dimensionality through downsampling (a feature that doesn't exist in NAT) and using a convolution over an attention block with fewer parameters.

Additionally, our motivation for stacking convolutions and DiNAT comes from the benefits of convolutions that have been understated by recent works. Prior work has noted that though attention based methods outperform, they mainly do on a very large scale \cite{dosovitskiy2021image}. Primarily, lack of convolutions loses the benefit of quick and easy translational equivariance and requiring larger datasets/training time to perform as expected or a use of regularization \cite{touvron2021training}. This becomes important in domains such as emotion recognition, as datasets are tedious to collect as they require expert annotators and a mix of annotators requires agreement in labeling, which can be sometimes subjective\footnote {Kollias \cite{kollias2019affwild2} notes that many frames get thrown away due to annotator disagreement, making emotion datasets sparse and small. Additionally, the process itself is much more tedious than common classification datasets.}.

\subsection{Causal Dilated Neighborhood Attention}

We extend the dilated neighborhood attention structure introduced by Hassani and Shi \cite{hassani2022dilated} (\cref{sec:DiNA}). Their Dilated Neighborhood Attention is modeled by attention weights $\mathbf{A}_{i}^{(k, \delta)}$ for a dilated by $\delta$ DiNA layer

\begin{equation}
	\mathbf{A}_{i}^{(k, \delta)} = \begin{bmatrix}Q_i K_{\rho_{1}^{\delta}{(i)}}^T + B_{(i,\rho_{1}^{\delta}{(i)})} \\Q_i K_{\rho_{2}^{\delta}{(i)}}^T + B_{(i,\rho_{2}^{\delta}{(i)})} \\ \vdots \\ Q_i K_{\rho_{k}^{\delta}{(i)}}^T + B_{(i,\rho_{k}^{\delta}{(i)})}\end{bmatrix}, 
	\label{eq:DiNAtenq}
\end{equation}
where $\rho_{j}^{\delta}{(i)}$ is $i$'s $j$-th nearest neighbor, $B_{(i,\rho_{k}^{\delta}{(i)})}$ is bias corresponding to two tokens $i$ and $j$, $Q$ and $K$ are query and key projections of $X$. The neighboring values of each neighborhood of size $k$ dilated by a value $\delta$ is

\begin{equation}
	\mathbf{V}_{i}^{(k, \delta)} = \begin{bmatrix}V_{\rho_{1}^{\delta}{(i)}}^T & V_{\rho_{2}^{\delta}{(i)}}^T & \hdots & V_{\rho_{k}^{\delta}{(i)}}^T\end{bmatrix}^T.
	\label{eq:DiNAtenv}
\end{equation}

For each sliding window, we can define the output of each pixel by 
\begin{equation}
	\text{DiNA}_{k}^{\delta}{(i)} = softmax\left(\frac{\mathbf{A}_{i}^{(k, \delta)}}{\sqrt{d_k}}\right) \mathbf{V}_{i}^{(k, \delta)}, 
	\label{eq:DiNAten}
\end{equation}
where $DiNA$ is applied to each element $i$, where $i \in \mathbb{R}^{1 \times n}$, and the output is then $DiNA(i)$. $V$, $Q$, and $K$ are all calculated in the same manner as self-attention, as $DiNA(i)$ tends to simply self-attention as you increase $k$ and decrease $d$ to 1.

In order to stop temporal leakage, we must ensure that for an input $(x_t | x_0,..., x_{t-1})$ to each temporal block, the output $(y_t | y_0,..., y_{t-1})$ must be influenced by a time step at least 1 less than $t$. To ensure this, we change $\rho_{j}^{\delta}{(i)}$ to represent the nearest neighbor to the left of the $i$-th value with a dilation $\delta$. This would mean that for a neighborhood $k$, the farthest value referenced is $\delta \cdot (k-1)$ to the left of the input sequence (not including the $x_t$ itself), rather than $\frac{\delta \cdot (k-1)}{2}$. In order to ensure this, we pad $\text{DiNA}_{k}^{\delta}{(i)}$ and the convolutions using casual padding in order to make sure that timesteps are not influenced by the future, then removing padding before the next temporal block to ensure length consistency. In implementation, this is a standard  $\delta \cdot (k-1) $ zero padding followed by removing $\delta \cdot (k-1)$ elements to the right, removing future timesteps. 

\subsection{Residual Connections}
\label{sec:residual}

Since a network requiring a large receptive field will require an increase in layers, a residual connection \cite{he2015deep} is added to address vanishing and shattering gradients problems \cite{zaeemzadeh2020normpreservation, balduzzi2018shattered, veit2016residual}, improve the loss landscape \cite{li2018visualizing} leading to more stable training and better results. Residual layers are simply described as 
\begin{align}
	H(x) = F(x) + x.
	\label{residual-1}
\end{align}
Since Temporal Blocks commonly upscale or downscale inputs, the residual layer in the NAC-TCN Temporal Block is 
\begin{align}
	H(x) = F(x) + G(x).
	\label{residual-1}
\end{align}
where $G(x)$ is an optional 1x1 convolution used when scaling of channels is required. The 1x1 convolution impact is twofold: reducing dimensions of the network in later layers and providing a way for the model to translate features from one layer to another while maintaining the same overall information as previous layers. DiNA is not used for this 1x1 convolution because of its inability perform dimension scaling.

NAC-TCN reduces parameters compared to the multi-model approach proposed by others \cite{Zhao2020-kw,DBLP:journals/corr/abs-2002-12530, zhang2022continuous}, the original TCN, and TCAN \cite{hao2020temporal}. This can be attributed to the fact that attention operations, which solely consider the kernel size of the neighborhood, are intertwined with the convolution operations, leading to a decrease in parameters when compared to traditional combined structures.

\section{Experiments}

In order to evaluate the effectiveness of our TCN methods, we used a variety of emotion and action recognition datasets, where newer temporal information is more relevant than the past. The $regnet\_y\_400mf$ image encoder \cite{9156494} is used as an encoder for all the datasets to ensure that the NAC-TCN is the main factor tested.

\paragraph{The AffWild2 dataset} \cite{kolakowska2014emotion,kollias2019deep,kollias2019expression, kollias2019face, kollias2020analysing, kollias2021affect, kollias2021analysing, kollias2021distribution, kollias2022abaw}  supplies 1,500,000+ annotated video frames of the valence and arousal metric in 341 videos.  A video length of 256 frames is used. Due to the fact that valence and arousal are between $[-1,1]$, $\tanh$ is applied to the model output. The valence and arousal scores are evaluated and trained on the Concordance Correlation Coefficient (CCC) metric
\begin{align}
	CCC={\frac {2s_{xy}}{s_{x}^{2}+s_{y}^{2}+({\bar {x}}-{\bar {y}})^{2}}},
	\label{eq:ccc}
\end{align}
where $x$ and $y$ are predictions and ground truths, ${2s_{x}}$ and ${2s_{y}}$ are variances, and ${\bar {x}}$ and ${\bar {y}}$ are the mean values.

\paragraph{The EmoReact dataset} \cite{nojavanasghari2016emoreact} provides videos of children annotated for 8 different emotions. Sequence length of 128 is used. The classes are not mutually exclusive and imbalanced. A random sampler and binary cross entropy are used to address these issues. The same CNN encoder model from the AffWild2 experiments is used. Evaluation is done using Area under the precision-recall curve (AUC-ROC) to follow similar methodology to prior studies. AUC-ROC tells us for different threshold how a model performs by ploting False Positive (FP) rate and True Positive (TP) rate and defining AUC-ROC as the area under this curve.

\paragraph{AFEW-VA} \cite{afew-1,afew-2} provides valence and arousal annotations to popular films. These annotations are integers ranging from [-10,10]. These are converted to mood labels to compare to prior works \cite{narayana2023focus}. Accuracy is used as the evaluation metric. The sequence length is 32, as videos are much shorter compared to other datasets. 

\subsection{Model Testing Methodology}

The baseline GRU and LSTM hyperparameters for the AffWild2 dataset are chosen to match the prior models tested on the dataset. Other models such as TCN \cite{BaiTCN2018} and TCAN \cite{hao2020temporal} used the same hyperparameters as the large NAC-TCN. GRU and LSTM blocks are concatenated with attention based models in order to create an ensemble of the two for testing. 

For each evaluated dataset, NAC-TCN was tested in two sizes. The larger size attempted to use similar hyperparameters to the GRU models while ensuring optimal receptive field \eqref{Equation1} through $k$, $d$, and number of layers. The optimal receptive field for all models besides AFEW-VA and AffWild2 were the length of the sequence, as only the final item was annotated. AFEW-VA used the entire sequence length $32$ and AffWild2 used $256$ based on prior literature. The model smaller size still ensured the optimal receptive field, but attempted to be a equal to smaller size than the GRU and LSTM models through adjusting previously mentioned hyperparameters along with the number of channels for the convolutional layers. This approach allowed us to conduct comparative tests while highlighting the versatility of the NAC-TCN model in terms of memory and computational cost.

\subsection{Implementation Details} 
We use an Adam Optimizer with a base learning of $0.001$ alongside an annealing cosine scheduler. We use a batch size of 16 for the AffWild2, EmoReact, and AFEW-VA datasets and all models were trained for 10 epochs. AFEW and AffWild2 all used subject based k-fold cross validation to ensure that information leakage did not occur between testing and training. The validation dataset of AffWild2 was used as the evaluation set and kept separate from training data. The EmoReact dataset had preset train and test splits that were used to be in line with the performance of prior models. The best performing model was selected for each method. The same random seed value was selected to ensure reproducibility. The number of heads for DiNAT was selected using hyperparameter search ($\{2^n \mid 1 \leq n \leq 3\}$). Note that for AffWild2, we build off the open source library provided by Nguyen 22 for testing our new model to ensure consistency when comparing to other methods \cite{Nguyen_2022_CVPR}.

\subsection{Metrics}

In addition to the per-dataset metrics, both operations and parameters are recorded. Operations are measured in MACs, or Multiply-Accumulate operations\footnote{Note that the FLOP operation is $2 \times \textit{MAC}$}. Both of these were measured using the Pytorch FLOPS Counter \cite{FlopsCounterPytorch}. $\text{MACs}$ represent the common operation of
\begin{align}
	(XW) + B,
	\label{gmac}
\end{align}
where $X$ is the original input, $W$ is a weight, and $B$ is a bias. We commonly report MMac, or MegaMACs ($10^6$ MACs).

\section{Results}

In this section, we report the performance of our proposed NAC-TCN architectures against the baseline GRU, LSTM, TCN, and attention models with both performance and efficiency. In addition, we compare against state-of-the-art models on the respective datasets where relevant.
\subsection{AffWild2}

\begin{table}[!htbp]
	\centering
	\caption{Results on AffWild2 Validation. All models reproduced besides competition provided baseline. Bold denotes indicates the highest performing model.}
	\begin{tabular}{@{}lclc}
		\toprule
		Method                                  & CCC $\uparrow$ & M Params  $\downarrow$ & MMac $\downarrow$ \\ \midrule
		Dataset Baseline \cite{kollias2022abaw} &      0.17      & --                     &        --         \\
		Attn.                                   &      0.20      & 0.97                   &        124        \\
		
		TCN                                     &      0.41      & 13.95                  &       3079        \\
		GRU \cite{Nguyen_2022_CVPR}             &      0.42      & 4.66                   &        597        \\
		LSTM                                    &      0.42      & 6.21                   &        797        \\
		GRU/Attn. \cite{Nguyen_2022_CVPR}       &      0.44      & 5.63                   &        721        \\
		LSTM/Attn                               &      0.44      & 6.79                   &        911        \\
		TCAN                                    &      0.46      & 17.12                  &       3700        \\
		Transformer \cite{chudasama2022m2fnet, vaswani2017attention} & 0.48 & 31.04& 3970 \\
		\midrule
		NAC-TCN (sm)                            & \textbf{0.48}  & \textbf{1.24  }                 &   \textbf{     381     }   \\
		NAC-TCN (lg)                            & \textbf{0.52}  & 10.12                  &       3230        \\ \bottomrule
	\end{tabular}
	
	\label{tab:full_affwild}
\end{table}

\begin{table*}[h!]
	\small
	\caption{Results on the EmoReact Dataset. Bold denotes indicates the highest performing model. Two variations of our proposed model architecture are reported, where sm indicates the smaller model size and lg is the larger model size. }
	\centering
	\begin{tabular}{lcccccc}
		\toprule
		Method                                                       & AUC ROC $\uparrow$ &    MParams $\downarrow$    & MMac $\downarrow$ & External Training Data & Audio  &   Video    \\ \midrule
		Attn.                                                        &        0.56        &      2.5       &        360        &                        &        &   \cmark   \\
		SVM \cite{nojavanasghari2016emoreact}                        &        0.62        &       --       &        --         &                        &        &   \cmark   \\
		SVM \cite{nojavanasghari2016emoreact}                        &        0.63        &       --       &        --         &                        & \cmark &   \cmark   \\
		GRU                                                          &        0.74        &      1.62      &        208        &                        &        &   \cmark   \\
		LSTM                                                         &        0.75        &      2.16      &        277        &                        &        &   \cmark   \\
		GRU/Attn.                                                    &        0.76        &      4.4       &        567        &                        & \cmark &            \\
		LSTM /Attn.                                                  &        0.76        &      4.9       &        636        &                        & \cmark &            \\
		TCN                                                          &        0.79        &      1.8       &        459        &                        &        &   \cmark   \\
		TSN  \cite{technologies9040086}    &        0.79        &       --       &        --         &         \cmark         &        &   \cmark   \\
		TCAN                                                         &        0.84        &      1.94      &        488        &                        &        &   \cmark   \\
		\midrule
		NAC-TCN (sm)                                                 &   \textbf{ 0.86}   & \textbf{ 1.5 } &        453        &                        &        &   \cmark   \\
		NAC-TCN (lg)                                                 &        0.78        &      7.8       &       2500        &                        &        &   \cmark   \\ \bottomrule
	\end{tabular}
	
	\label{tab:EmoReact}
\end{table*}

As reported in \cref{tab:full_affwild}, our proposed NAC-TCN architecture outperforms the other temporal-based models, while using a smaller model size. Additionally, we achieve the highest performance at smaller model sizes. The NAC-TCN, achieved through either a simple single layer setup or with self-attention, exhibited higher performance than the base TCN, which indicates that NAC-TCN can better learn temporal representations with less memory. 

It should be noted that superior performance has exhibited in recent studies. However, conducting a direct comparison is challenging due to the utilization of disparate datasets and the employment of multi-sensor methodologies during the training process. Notably, participants in the latest AffWild2 challenge have surpassed reported results through four-encoder model, which incorporates audio and image encoders \cite{meng2022multi}. Other participants have surpassed prior state-of-the-art results with use of linguistic models from extracted words \cite{zhang2023multimodal}. For our purposes, we have achieved a state-of-the-art result in the chosen set of input modalities and encoder choice.



\subsection{EmoReact}

Results on EmoReact \cite{nojavanasghari2016emoreact} (\cref{tab:EmoReact}) show that with less modalities, NAC-TCN Small outperforms other models \emph{without} multiple modalities or increased training data. This is done with a \textit{decrease} in parameters and operations. This indicates that a better performing architecture like NAC-TCN may be actually outperform even with less data. NAC-TCN may be more prone to overfitting, given that with similar parameters to GRU and LSTM, it performed similarly and worse to a larger TCN. This highlights that NAC-TCN can be more expressive with the same hyperparamters, hence strong of the smaller model. 

\subsection{AFEW-VA}

\begin{table}[!h]
	\centering
	\caption{Results on AFEW-VA dataset mood labels. Note that 1-CNN used a student teacher learning paradigm which may increase runtime beyond what is expected. }
	\begin{tabular}{lccc}
		\toprule
		Method                                     &   Accuracy $\uparrow$    & K Params $\downarrow$ & MMac $\downarrow$ \\ \midrule
		Attn.                                       &     0.43 $\pm$ 0.12      &              1860         &   234                \\
		1-CNN  \cite{narayana2023focus}            &     0.70 $\pm$ 0.10      &          --           &        --         \\
		TS (Mood/$\Delta$)\cite{narayana2023focus} &     0.73 $\pm$ 0.08      &          --           &        --         \\
		TCN                                        &     0.74 $\pm$ 0.05      &         1220          &        99         \\
		GRU                                        &     0.75 $\pm$  0.05     &          374          &        10         \\
		LSTM                                       &     0.75 $\pm$ 0.04      &          423          &        13         \\
		TCAN                                        &     0.75 $\pm$ 0.06      &         1253         &        120         \\
		LSTM/Attn                                  &     0.75 $\pm$ 0.46      &         1260          &        40         \\
		GRU/Attn                                   & \textbf{0.76 $\pm$ 0.14} &         1150          &        36         \\ \midrule
		NAC-TCN (lg)                                   & \textbf{0.76 $\pm$ 0.12} &          988          &        84         \\
		NAC-TCN (sm)                               &     0.75 $\pm$ 0.09      &          204          &        17         \\ \bottomrule
	\end{tabular}
	
	\label{tab: AFEW}
\end{table}

AFEW-VA results show that the larger NAC-TCN is able to outperform other methods. The smaller model results in similar performance to TCNs but with smaller memory footprint.  It is important to note that the disparity between the models is minimal, within a $\pm$ 2\% range. Consequently, the AFEW-VA dataset should be regarded primarily as a validation of the NAC-TCN's capacity to maintain performance levels akin to those of more expansive models. Nevertheless, NAC-TCN outperforms the 1-CNN model which uses attention \cite{narayana2023focus}.

\subsection{Ablation Studies}

In order to understand the impact of different choices we made in design and experimentation, we perform several ablation studies. 

\subsubsection{Residual Connection}

\begin{table}[!htbp]
	\centering
	\small
	\caption{Ablation study comparing residual connection on NAC-TCN small on the AffWild2 dataset. }
	\begin{tabular}{@{}lcc}
		\toprule
		CCC $\uparrow$ &  Residual  &  \\ \midrule
		0.41              & \xmark &  \\
		0.48           & \cmark &  \\
		\bottomrule
	\end{tabular}
	
	\label{tab:affwild2_res}
\end{table}
We conduct on ablation study on the NAC-TCN Small model with the AffWild2 dataset, since the dataset has the deepest model due to the larger receptive field.

The study reveals that the residual connection is critical to training. Without residual connection, the model saw a significant performance decrease (\cref{tab:affwild2_res}).

\subsubsection{Importance of Casuality}
\begin{table}[!hb]
	\centering
	\small
	\caption{Small model used for NAC-TCN. Recall to \cref{tab:EmoReact} and \cref{tab:full_affwild}.}
	\begin{tabular}{ccccc}
		\toprule
		Model                 &   Causal   & \multicolumn{1}{c}{Affwild2} & \multicolumn{1}{c}{EmoReact} &  \\
		\cmidrule(lr){3-3} \cmidrule(lr){4-4} &            &             CCC              &           AOC-ROC            &  \\ \midrule
		NAC-TCN                & \cmark &             0.48             &             0.86             &  \\
		NAC-TCN                &   \xmark   &             0.44             &             0.65             &  \\ \bottomrule
	\end{tabular}
	
	\label{tab:causal}
\end{table}
We compare the robustness to causality of our model versus other similar models in \cref{tab:causal}. Our model weights attention and applies convolutions based on previous $k$ timesteps, where an acausal model would weight based on $\frac{k}{2}$ on each side. We find that the causal relation is important in both datasets, but is more dramatic in the EmoReact dataset. This suggests that emotions are better learned when future information is unknown. This phenomenon of better learning with less information can be attributed to two potential reasons. Firstly, emotions inherently involve a causal process, wherein per-frame annotations occur continuously, thereby influencing annotators' decisions based on prior frames rather than knowledge of future frames. This can lead to different understandings depending on what context is used. Secondly, the disparity between the datasets stems from the variation in label format. AffWild2 employs per-frame labels, allowing for non-causal predictions of adjacent frames, whereas EmoReact utilizes end-of-video labels, thereby elevating the significance of causality (the last frame culminating in information from previous frames rather than $h$/2 prior frames). We find that prior literature commonly uses causal relationships over acausal with better results, making it an interesting point of discussion for future work.  

\section{Discussion}

\subsection{Limitations}
Although our method outperformed on many datasets, performance on AFEW-VA is notably similar to other temporal models. Given AFEW-VA is a smaller dataset, this may indicate that NAC-TCN outperforms other models on larger datasets with more oracle access. Multi-model \& pretraining approaches that could perform better were not studied due to hardware limitations and simplicity in results. Our model also holds many of the same flaws of modern TCN based methods, such as higher memory during evaluation (needing the whole sequence instead of hidden state) and poor transfer learning with different $k$ or $d$ values. 
\subsection{Contribution}
In this paper, we presented an alternative to the Temporal Convolutional Network that allows for attention while \textit{decreasing parameters and number of MAC operations}. Experimental evaluation revealed improvements over classical methods such as GRUs, LSTMs, and Attention-based methods at a \textit{lower computational cost}. Our method outperforms common temporal methods, improves on the benefits of the TCN, and performs similarly at an efficiency benefit while maintaining the same TCN controls over memory usage.


\bibliographystyle{ACM-Reference-Format}
\bibliography{acmart}

\end{document}